\documentclass{article}
\usepackage{soul}
\usepackage[dvipsnames]{xcolor}

    \usepackage[final, nonatbib]{neurips_2020}

\usepackage[utf8]{inputenc} 
\usepackage[T1]{fontenc}    
\usepackage{hyperref}       
\usepackage{url}            
\usepackage{booktabs}       
\usepackage{amsfonts}       
\usepackage{nicefrac}       
\usepackage{microtype}      
\usepackage{xcolor}
\usepackage{graphicx}
\usepackage{capt-of}
\usepackage{amsmath}
\usepackage{scalerel}
\usepackage{tablefootnote}
\usepackage{subfig}

\newcommand{\Lagr}{\mathcal{L}}

\title{End-to-End Differentiable 6DoF Object Pose Estimation with Local and Global Constraints}

\author{%
  Anshul Gupta, Joydeep Medhi, Aratrik Chattopadhyay, Vikram Gupta \\
  Mercedes-Benz Research and Development India\\
  \texttt{firstname.lastname@daimler.com}
}

\begin{document}

\maketitle

\begin{abstract}
Inferring the 6DoF pose of an object from a single RGB image is an important but challenging task, especially under heavy occlusion. While recent approaches improve upon the two stage approaches by training an end-to-end pipeline, they do not leverage local and global constraints. 
In this paper, we propose pairwise feature extraction to integrate local constraints, and triplet regularization to integrate global constraints for improved 6DoF object pose estimation.
Coupled with better augmentation, our approach achieves state of the art results on the challenging Occlusion Linemod dataset, with a $9\%$ improvement over the previous state of the art, and achieves competitive results on the Linemod dataset.
\end{abstract}

\section{Introduction}

Estimating the 6DoF pose of an object is an important problem with applications in various domains like robotics~\cite{correll2016analysis}, augmented reality~\cite{arpose} and autonomous driving~\cite{song2019apollocar3d}. With the pervasion of inexpensive RGB sensors, it is cost effective and highly beneficial to perform 6DoF pose estimation from a single RGB image without using additional depth sensors.

Some studies~\cite{kehl2017ssd}\cite{xiang2018posecnn} attempted to regress the 6DoF pose directly from the image, however, these were not as competitive as recent two stage approaches. In the first stage of two stage approaches, a \textit{correspondence estimator} detects the object and estimates the 2D image projections of the 3D object points (referred to as 2D keypoints). 
This establishes correspondences between the 2D and 3D points. \cite{rad2017bb8}\cite{tekin2018real}\cite{hu2019segmentation} used a CNN based architecture to segment out regions containing the object and regress the 2D keypoints from those regions. A recent study regressed direction vectors to the 2D keypoints from the segmented regions of the object~\cite{peng2019pvnet}. The 2D keypoints were then estimated from intersections of pairs of direction vectors. This approach was found to be more robust to occlusions of the object.


In the second stage, a RANSAC based Perspective-n-Point(PnP) algorithm serves as a \textit{pose estimator} to predict the 6DoF object pose using the established 2D-3D correspondences. However, Hu et al.~\cite{hu2020single} showed that RANSAC is sensitive to the ordering of the 2D-3D correspondences and computationally costly when there are many of them. Further, the non-differentiable nature of the RANSAC based pose estimator 
does not allow for end-to-end training of the two stage approaches with respect to the final objective, namely the object pose. Hence, Hu et al.~\cite{hu2020single} proposed to replace the non-differentiable RANSAC based pose estimator with a trainable neural network to estimate the 6DoF object pose. Their end-to-end trainable model showed improved results compared to the two stage approach as validated with two state of the art correspondence estimators~\cite{hu2019segmentation}~\cite{peng2019pvnet}. We follow up on their model with~\cite{peng2019pvnet} as the correspondence estimator as it shows superior performance and refer to it as SSPE.

     

While SSPE shows improved performance using end-to-end training, it does not utilize local and global geometric constraints. In this work, we propose pairwise features to utilize local information between direction vectors associated with the same 3D point, and triplet regularization to account for the global geometry between pairwise features associated with different 3D points.
 Coupled with increased masking augmentation, our model achieves state of the art results on the Occlusion Linemod~\cite{occlusionlinemod} dataset and competitive results on the Linemod~\cite{linemod} dataset. In summary, our main contributions are:
\begin{itemize}

    \item Pairwise feature extraction from direction vectors to better utilize local information
    \item Triplet regularization to account for the global geometry of the pairwise features
    \item State of the art results on Occlusion Linemod and competitive results on Linemod 

\end{itemize}

\section{Approach}

\begin{figure*}[t]
\centering
\includegraphics[width=1\linewidth]{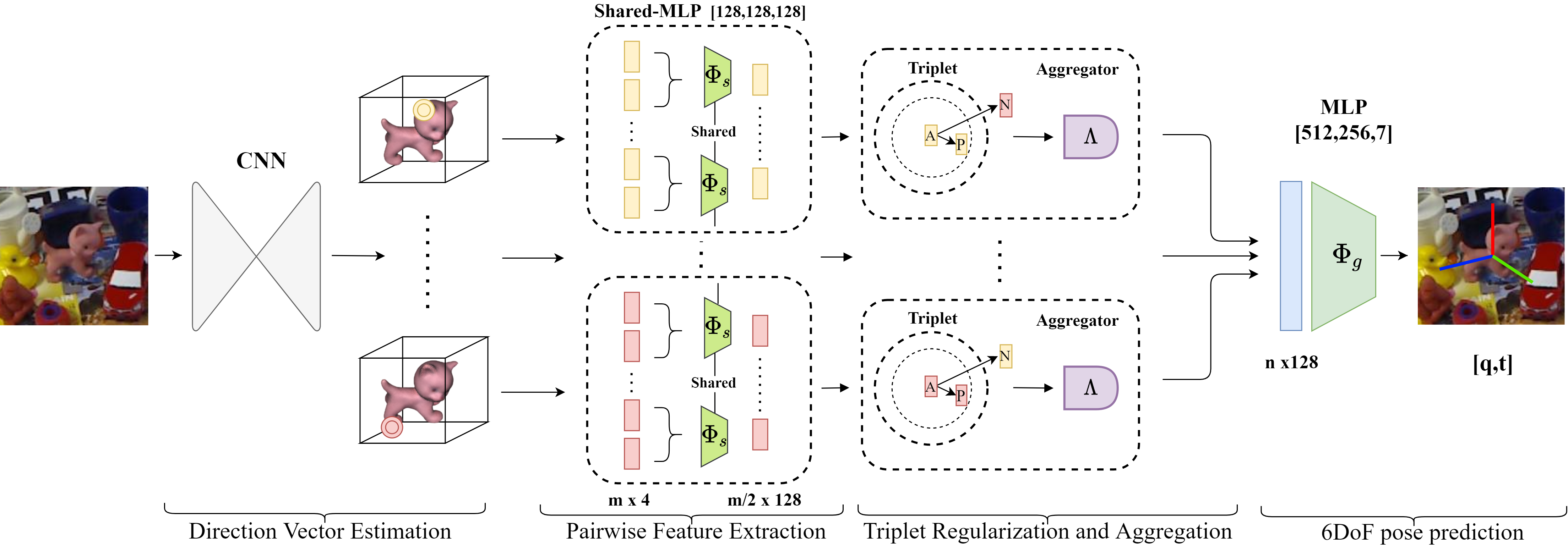}
\caption{Illustration of our network architecture (SSPE-ours). The correspondence estimator predicts direction vectors to the 2D keypoints. Pairs of direction vectors are passed through a shared network $\Phi_s$ to give pairwise features which are aggregated using an aggregator $\Lambda$, and passed through a second network $\Phi_g$ to predict the pose. The color of the pairwise features indicates association to a 3D point.}
\label{fig:diagram}
\end{figure*}

We illustrate our approach in Figure~\ref{fig:diagram}. The correspondence estimator operates on an image and predicts a segmentation mask. It also predicts direction vectors to the 2D keypoints for each pixel in the mask. For each of the $n$ 3D points $p_i$, the pose estimator selects $m$ random direction vectors $u_{ik}$ ($1 \leq i \leq n$, $1 \leq k \leq m$) from the segmented region of the object. It applies a shared MLP $\Phi_s$ to extract pairwise features, followed by aggregation using an aggregator $\Lambda$, and pose prediction from a second MLP $\Phi_g$.


\subsection{Local Constraint}


A direction vector $u_{ik}$ is represented as a 4D input $[x, y, dx, dy]$ where $x,y$ is the pixel location, and $dx, dy$ is the predicted vector from that pixel. In the first step for the SSPE pose estimator, a shared MLP is applied across all direction vectors to extract $n \times m$ local features. 
However, 
by operating on every direction vector independently the local features do not have information about the 2D keypoints. This is because a 2D keypoint is given by the intersection of a pair of direction vectors pointing to that keypoint~\cite{peng2019pvnet}. Hence, we propose to concatenate pairs of direction vectors $[u_{ik}, u_{il}]$ and provide them as input to the shared MLP $\Phi_s$. This gives us $n \times \frac{m}{2}$ $D$ dimensional
features $f_{ih}$ ($1 \leq h \leq \frac{m}{2}$) termed as \textit{pairwise features}.
\begin{equation}
f_{ih} = \Phi_s([u_{ik}, u_{il}]) \hspace{1cm} \scaleto{1 \leq i \leq n, \hspace{0.2cm} 1 \leq h \leq \frac{m}{2}, \hspace{0.2cm} k=2h-1, \hspace{0.2cm} l=2h}{12pt}
\end{equation}

While $\Phi_s$ can theoretically learn to approximate the intersection of direction vectors to give pairwise features with information about the 2D keypoints, we observe that adding global constraints can help learn better features for improved performance.



\subsection{Global Constraint}

\begin{table}[t]
\centering
\caption{Results on Occlusion Linemod (Part I) and Linemod (Part II) using the ADD0.1d metric$^1$.}
\label{tab:results-occlinemod}
\centering 
\scriptsize
\begin{tabular}{ c | c c c c c | c c c c }
\hline
& \multicolumn{5}{c|}{\textbf{Part I: Occlusion Linemod}} &  \multicolumn{4}{c}{\textbf{Part II: Linemod}}\\
& \textbf{PVNet}~\cite{peng2019pvnet} & \textbf{DPVR}~\cite{proxyvoting} & \textbf{SSPE}~\cite{hu2020single} & \textbf{SSPE-r}$^2$ & \textbf{SSPE-ours} & \textbf{PVNet}~\cite{peng2019pvnet} & \textbf{DPVR}~\cite{proxyvoting} & \textbf{SSPE-r} & \textbf{SSPE-ours}\\
\hline
Ape & 15.8 & 19.2 & 19.2 & \textbf{20.8} & 18.8 & 43.6 & \textbf{69.1} & 66.7 & 52.5\\
Can & 63.3 & 69.8 & 65.1 & 78.4 & \textbf{79.3} & 95.5 & 98.5 & 95.8 & \textbf{99.2}\\
Cat & 16.7 & \textbf{21.1} & 18.9 & 18.2 & 17.5 & 79.3 & 83.1 & 84.1 & \textbf{88.5}\\
Driller & 65.7 & 71.6 & 69.0 & 73.8 & \textbf{76.4} & 96.4 & \textbf{99.0} & 98.4 & 98.8\\
Duck & 25.2 & 34.3 & 25.3 & 33.1 & \textbf{34.4} & 52.6 & 63.5 & 60.4 & \textbf{68.7}\\
Eggbox* & 50.2 & 47.3 & \textbf{52.0} & 46.0 & 44.6 & 99.2 & \textbf{100.0} & 99.7 & \textbf{100.0}\\
Glue* & 49.6 & 39.7 & 51.4 & 49.2 & \textbf{53.2} & 95.7 & 98.0 & 90.4 & \textbf{98.5}\\
Holepuncher & 39.7 & 45.3 & 45.6 & 53.5 & \textbf{54.7} & 81.9 & \textbf{88.2} & 85.3 & 88.1\\
\hline
Average & 40.8 & 43.5 & 43.3 & 46.6 & \textbf{47.4} & 80.5 & \textbf{87.4} & 85.1 & 86.8\\
\hline

\multicolumn{10}{l}{$^1$We do not compare against models that perform refinement on predicted pose~\cite{li2018deepim}\cite{wang2019densefusion}.}\\
\multicolumn{10}{l}{$^2$We reimplement SSPE as authors have not open sourced the training code}
\end{tabular}
\end{table}


We account for the global geometry of the pairwise features by considering their association to the 3D points. We want pairwise features associated with the same 3D point to be similar to each other, and pairwise features associated with different 3D points to be dissimilar to each other. To encourage this property we introduce a triplet regularization term. This also serves as a form of proxy supervision to the shared MLP $\Phi_s$ as different pairs of direction vectors associated with the same 3D point give similar pairwise features. 
We mine triplets online and compute the triplet regularization term as:
\begin{equation}
\Lagr_t = \frac{2}{nm} \sum_{i=1}^{n} \sum_{h=1}^{\frac{m}{2}} max(S_{ih,jd} - S_{ih,is} + \alpha, 0) \hspace{1cm} \scaleto{1\leq j \leq n, \hspace{0.2cm} i\neq j, \hspace{0.2cm} 1 \leq d,s \leq \frac{m}{2}}{12pt}
\end{equation}
where $\alpha$ is the margin and $S_{wx,yz}$ is the similarity between pairwise features $f_{wx}$ and $f_{yz}$. We use the cosine similarity function given as:
\begin{equation}
S_{wx,yz} = \frac{f_{wx}^{T}f_{yz}}{||f_{wx}|| \hspace{0.1cm}||f_{yz}||} \hspace{1cm} \scaleto{1\leq w,y \leq n, \hspace{0.2cm} 1 \leq x,z \leq \frac{m}{2}}{12pt}
\end{equation}

Similar to SSPE, we aggregate the pairwise features and apply a second MLP to compute the pose. 
The pairwise features associated with each 3D point are aggregated using an aggregator $\Lambda$ to give $n$ $D$ dimensional group features $g_i$. We choose $\Lambda$ as the mean pooling aggregator. 
\begin{equation}
g_i = \Lambda(\{f_{i1}, f_{i2} ... f_{i\frac{m}{2}}\}) \hspace{1cm} \scaleto{1 \leq i \leq n}{6pt}
\end{equation}
The group features are concatenated, and the $nD$ dimensional vector is passed through a second MLP $\Phi_g$ to predict the pose as a quaternion $\hat{q}$ and translation $\hat{t}$. 
\begin{equation}
    [\hat{q},\hat{t}] = \Phi_g([g_1, g_2 ... g_n])
\end{equation}

We recover the predicted rotation matrix $\hat{R}$ from $\hat{q}$ and compute the pose loss $\Lagr_p$ as the 3D error:
\begin{equation}
\Lagr_p = \frac{1}{n} \sum_{i=1}^{n} ||(\hat{R}p_i + \hat{t}) - (Rp_i + t) || 
\end{equation}
where $R$ and $t$ are the ground truth rotation and translation.

The final loss $\Lagr$ to optimize is a linear combination of the cross entropy segmentation loss $\Lagr_s$ and L1 vector regression loss $\Lagr_k$ from the correspondence estimator~\cite{peng2019pvnet}, and the pose loss $\Lagr_p$ and triplet regularization term $\Lagr_t$ from the pose estimator.
\begin{equation}
\Lagr = \lambda_s\Lagr_s + \lambda_k\Lagr_k + \lambda_p\Lagr_p + \lambda_t\Lagr_t
\end{equation}

\section{Experiments}




\subsection{Training}

We use $n=9$ 3D key points for each object selected using the farthest point sampling algorithm. For the pose estimator, we randomly select $m=200$ direction vectors for each of the 3D points. The triplet margin $\alpha$ is set to $0.1$. The loss coefficients $\lambda_s$ and $\lambda_k$ are set to 1, $\lambda_p$ is set to $0.01$ and $\lambda_t$ is set to 0.1. As per previous studies~\cite{peng2019pvnet}\cite{hu2020single}, we train separate models for each object. Training images are provided at an input resolution of $640 \times 480$ and augmented using scaling, translation, rotation, occlusion~\cite{zhong2020random}, gaussian blurring and colour jittering. We use the Adam optimizer and set the learning rate to $1e-3$ which is divided by 10 after processing $50\%$, $75\%$, and $90\%$ of the data. All models are trained with a batch size of $32$ for $300$ epochs.

\subsection{Evaluation}
We benchmark our approach on the Linemod~\cite{linemod} and Occlusion Linemod~\cite{occlusionlinemod} datasets for 8 object classes. Similar to previous approaches~\cite{peng2019pvnet}\cite{hu2020single}, we augment the Linemod train data using synthetic data. We generate $10000$ images containing multiple objects using the cut and paste~\cite{dwibedi2017cut} technique, and $8\times10000$ images of single objects using the rendering technique in~\cite{peng2019pvnet}. 

For evaluation, we use the ADD0.1d metric~\cite{linemod} to measure accuracy in 3D space. 
The ADD0.1d metric measures the average distance between the 3D model points transformed using the predicted pose and the ground truth pose. A predicted pose is assumed correct if the average distance is less than $10\%$ of the model diameter. We report the percentage of correctly predicted poses. We use the symmetric version of the metric~\cite{xiang2018posecnn} for symmetric objects, which are denoted by the * superscript.


\subsection{Results}

\begin{figure}[t]
\begin{minipage}{0.4\linewidth}
  \centering
    \subfloat[]{
    \includegraphics[width=0.465\linewidth]{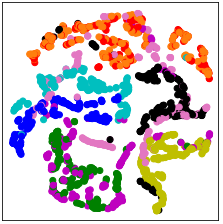}
    }
    \subfloat[]{
    \includegraphics[width=0.48\linewidth]{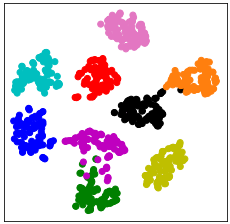}
    }

  \caption{t-SNE plot of the SSPE-r local features (a) and the SSPE-ours pairwise features (b) for the holepuncher object. Each colour represents the features of the $9$ 3D points.}
  \label{fig:tsne}
\end{minipage}
\hspace{0.4cm}
\begin{minipage}{0.55\linewidth}
\centering
\scriptsize
\captionof{table}{Ablation study on $4$ non-symmetric and $1$ symmetric object. Adding local constraints (SSPE-lc) improves performance over SSPE-r and SSPE-rp. Adding global constraints (SSPE-ours) further improves performance. Results reported on Occlusion Linemod using the ADD0.1d metric.}
\label{tab:ablation}
\begin{tabular}{ c | c c c c c}
\hline
\hspace{0.1cm}
& \textbf{SSPE-m} & \textbf{SSPE-r} & \textbf{SSPE-rp} & \textbf{SSPE-lc} & \textbf{SSPE-ours}\\
\hline
Can & 71.9 & 78.4 & 79.5 & 77.6 & 79.3\\
Driller & 62.4 & 73.8 & 75.5 & 76.1 & 76.4\\
Duck & 28.8 & 33.1 & 32.3 & 32.3 & 34.4\\
Glue* & 51.3 & 49.2 & 53.6 & 55.9 & 53.2\\
Holepuncher & 44.7 & 53.5 & 51.1 & 52.7 & 54.7\\
\hline
Average & 51.8 & 57.6 & 58.4 & 58.9 & 59.6\\
\hline
\end{tabular}
\end{minipage}
\end{figure}

We report results on the Occlusion Linemod dataset in Part I of Table~\ref{tab:results-occlinemod}. SSPE-ours achieves state of the art results with a $9\%$ improvement over the previous best method~\cite{proxyvoting}. It has the highest scores for $5$ of the $8$ objects. 

We perform ablation in Table~\ref{tab:ablation} to demonstrate the strength of our approach. Average performance of SSPE with pairwise features (SSPE-lc) is better compared to standard SSPE with the aggregator as max pooling (SSPE-r) and SSPE with the aggregator as mean pooling (SSPE-rp). Adding triplet regularization (SSPE-ours) further improves performance. To support our hypothesis we do a t-SNE visualisation of the SSPE local features and our pairwise features as shown in Figure~\ref{fig:tsne}. We note much better clustering for our pairwise features. This suggests our approach successfully accounts for the local and global constraints to improve end-to-end pose estimation.

We also observe that increased masking augmentation~\cite{zhong2020random} can help increase performance. We highlight its importance in Table~\ref{tab:ablation} by initially setting the masking percentage to $10\%-30\%$ (SSPE-m), and then tripling it to $30\%-90\%$ (SSPE-r). We note an average increase of $5.8$ points in ADD0.1d score. Hence, we use the increased masking in all our experiments.

We additionally show results on the Linemod dataset in Part II of Table~\ref{tab:results-occlinemod}. SSPE-ours achieves competitive results and has the highest scores for $5$ of the $8$ objects. It also shows improvement over SSPE-r.

\section{Conclusion}

We show that our approach (SSPE-ours) achieves state of the art results on the challenging Occlusion Linemod dataset. We also perform ablation to demonstrate the strength of our approach. This suggests the effectiveness of local and global constraints to improve end-to-end 6DoF object pose estimation. In the future, we hope to explore geometric properties to further improve end-to-end 6DoF object pose estimation.

\bibliographystyle{IEEEtran}
\bibliography{mybib}

\begin{thebibliography}{10}
\providecommand{\url}[1]{#1}
\csname url@samestyle\endcsname
\providecommand{\newblock}{\relax}
\providecommand{\bibinfo}[2]{#2}
\providecommand{\BIBentrySTDinterwordspacing}{\spaceskip=0pt\relax}
\providecommand{\BIBentryALTinterwordstretchfactor}{4}
\providecommand{\BIBentryALTinterwordspacing}{\spaceskip=\fontdimen2\font plus
\BIBentryALTinterwordstretchfactor\fontdimen3\font minus
  \fontdimen4\font\relax}
\providecommand{\BIBforeignlanguage}[2]{{%
\expandafter\ifx\csname l@#1\endcsname\relax
\typeout{** WARNING: IEEEtran.bst: No hyphenation pattern has been}%
\typeout{** loaded for the language `#1'. Using the pattern for}%
\typeout{** the default language instead.}%
\else
\language=\csname l@#1\endcsname
\fi
#2}}
\providecommand{\BIBdecl}{\relax}
\BIBdecl

\bibitem{correll2016analysis}
N.~Correll, K.~E. Bekris, D.~Berenson, O.~Brock, A.~Causo, K.~Hauser, K.~Okada,
  A.~Rodriguez, J.~M. Romano, and P.~R. Wurman, ``Analysis and observations
  from the first amazon picking challenge,'' \emph{IEEE Transactions on
  Automation Science and Engineering}, vol.~15, no.~1, pp. 172--188, 2016.

\bibitem{arpose}
E.~{Marchand}, H.~{Uchiyama}, and F.~{Spindler}, ``Pose estimation for
  augmented reality: A hands-on survey,'' \emph{IEEE Transactions on
  Visualization and Computer Graphics}, vol.~22, no.~12, pp. 2633--2651, 2016.

\bibitem{song2019apollocar3d}
X.~Song, P.~Wang, D.~Zhou, R.~Zhu, C.~Guan, Y.~Dai, H.~Su, H.~Li, and R.~Yang,
  ``Apollocar3d: A large 3d car instance understanding benchmark for autonomous
  driving,'' in \emph{Proceedings of the IEEE Conference on Computer Vision and
  Pattern Recognition}, 2019, pp. 5452--5462.

\bibitem{kehl2017ssd}
W.~Kehl, F.~Manhardt, F.~Tombari, S.~Ilic, and N.~Navab, ``Ssd-6d: Making
  rgb-based 3d detection and 6d pose estimation great again,'' in
  \emph{Proceedings of the IEEE International Conference on Computer Vision},
  2017, pp. 1521--1529.

\bibitem{xiang2018posecnn}
Y.~Xiang, T.~Schmidt, V.~Narayanan, and D.~Fox, ``Posecnn: A convolutional
  neural network for 6d object pose estimation in cluttered scenes,'' 2018.

\bibitem{rad2017bb8}
M.~Rad and V.~Lepetit, ``Bb8: A scalable, accurate, robust to partial occlusion
  method for predicting the 3d poses of challenging objects without using
  depth,'' in \emph{Proceedings of the IEEE International Conference on
  Computer Vision}, 2017, pp. 3828--3836.

\bibitem{tekin2018real}
B.~Tekin, S.~N. Sinha, and P.~Fua, ``Real-time seamless single shot 6d object
  pose prediction,'' in \emph{Proceedings of the IEEE Conference on Computer
  Vision and Pattern Recognition}, 2018, pp. 292--301.

\bibitem{hu2019segmentation}
Y.~Hu, J.~Hugonot, P.~Fua, and M.~Salzmann, ``Segmentation-driven 6d object
  pose estimation,'' in \emph{Proceedings of the IEEE Conference on Computer
  Vision and Pattern Recognition}, 2019, pp. 3385--3394.

\bibitem{peng2019pvnet}
S.~Peng, Y.~Liu, Q.~Huang, X.~Zhou, and H.~Bao, ``Pvnet: Pixel-wise voting
  network for 6dof pose estimation,'' in \emph{Proceedings of the IEEE
  Conference on Computer Vision and Pattern Recognition}, 2019, pp. 4561--4570.

\bibitem{hu2020single}
Y.~Hu, P.~Fua, W.~Wang, and M.~Salzmann, ``Single-stage 6d object pose
  estimation,'' in \emph{Proceedings of the IEEE/CVF Conference on Computer
  Vision and Pattern Recognition}, 2020, pp. 2930--2939.

\bibitem{occlusionlinemod}
E.~Brachmann, A.~Krull, F.~Michel, S.~Gumhold, J.~Shotton, and C.~Rother,
  ``Learning 6d object pose estimation using 3d object coordinates,'' in
  \emph{European conference on computer vision}.\hskip 1em plus 0.5em minus
  0.4em\relax Springer, 2014, pp. 536--551.

\bibitem{linemod}
S.~Hinterstoisser, V.~Lepetit, S.~Ilic, S.~Holzer, G.~Bradski, K.~Konolige, and
  N.~Navab, ``Model based training, detection and pose estimation of
  texture-less 3d objects in heavily cluttered scenes,'' in \emph{Asian
  Conference on Computer Vision}.\hskip 1em plus 0.5em minus 0.4em\relax
  Springer, 2012, pp. 548--562.

\bibitem{proxyvoting}
X.~Yu, Z.~Zhuang, P.~Koniusz, and H.~Li, ``6dof object pose estimation via
  differentiable proxy voting loss,'' \emph{Proceedings of the British Machine
  Vision Conference}, 2020.

\bibitem{li2018deepim}
Y.~Li, G.~Wang, X.~Ji, Y.~Xiang, and D.~Fox, ``Deepim: Deep iterative matching
  for 6d pose estimation,'' in \emph{Proceedings of the European Conference on
  Computer Vision (ECCV)}, 2018, pp. 683--698.

\bibitem{wang2019densefusion}
C.~Wang, D.~Xu, Y.~Zhu, R.~Mart{\'\i}n-Mart{\'\i}n, C.~Lu, L.~Fei-Fei, and
  S.~Savarese, ``Densefusion: 6d object pose estimation by iterative dense
  fusion,'' in \emph{Proceedings of the IEEE Conference on Computer Vision and
  Pattern Recognition}, 2019, pp. 3343--3352.

\bibitem{zhong2020random}
Z.~Zhong, L.~Zheng, G.~Kang, S.~Li, and Y.~Yang, ``Random erasing data
  augmentation,'' in \emph{Proceedings of the AAAI Conference on Artificial
  Intelligence (AAAI)}, 2020.

\bibitem{dwibedi2017cut}
D.~Dwibedi, I.~Misra, and M.~Hebert, ``Cut, paste and learn: Surprisingly easy
  synthesis for instance detection,'' in \emph{Proceedings of the IEEE
  International Conference on Computer Vision}, 2017, pp. 1301--1310.

\end{thebibliography}

\end{document}